\documentclass[times,review,10pt]{elsarticle}

\usepackage{amsmath,amssymb,amsfonts}
\usepackage{booktabs,multirow,multicol,arydshln,colortbl}
\usepackage[table]{xcolor}
\usepackage{csquotes}
\usepackage{algorithm}
\usepackage{algorithmic}
\usepackage{microtype}
\usepackage{pifont}
\usepackage{adjustbox}
\usepackage{graphicx}
\usepackage{subcaption}
\usepackage{placeins}
\usepackage[colorlinks=true,linkcolor=blue,citecolor=blue,urlcolor=blue]{hyperref}

\journal{Pattern Recognition}

\begin{document}

\begin{frontmatter}

\title{Bi-Refine: Semantic Verification-Guided Progressive Refinement for Moment Retrieval}

\author[aff1,aff2]{Mengnan Liu}
\ead{Liumn@stu.xjtu.edu.cn}
\author[aff1,aff2]{Sanping Zhou\corref{cor1}}
\ead{spzhou@xjtu.edu.cn}
\author[aff3]{Kun Xia}
\ead{kunxia@xjtu.edu.cn}
\author[aff1,aff2]{Jiaxuan Lin}
\ead{linjiaxuan@stu.xjtu.edu.cn}
\author[aff1,aff2]{Yuxuan Wu}
\ead{yuxuan9862@gmail.com}
\author[aff4]{Gang Hua}
\ead{ganghua@gmail.com}

\cortext[cor1]{Corresponding author.}

\affiliation[aff1]{organization={National Key Laboratory of Human-Machine Hybrid Augmented Intelligence},
  addressline={Xi'an Jiaotong University},
  city={Xi'an},
  postcode={710049},
  state={Shaanxi},
  country={China}}
\affiliation[aff2]{organization={School of Artificial Intelligence},
  addressline={Xi'an Jiaotong University},
  city={Xi'an},
  postcode={710049},
  state={Shaanxi},
  country={China}}
\affiliation[aff3]{organization={School of Computer Science and Technology},
  addressline={Xi'an Jiaotong University},
  city={Xi'an},
  postcode={710049},
  state={Shaanxi},
  country={China}}
\affiliation[aff4]{organization={Amazon Alexa AI},
  city={Bellevue},
  postcode={98004},
  state={WA},
  country={USA}}

\begin{abstract}
Moment Retrieval (MR) aims to identify the temporal segment in an untrimmed video that corresponds to a natural language query. Existing approaches commonly refine temporal predictions within the localization pathway, without explicitly verifying candidate-level semantic consistency. Moreover, directly reusing semantic scores as feedback can cause over-correction when the verification signal is unreliable. To address these issues, we propose Bi-Refine, a semantic verification-guided progressive refinement framework that formulates MR as a closed-loop verification-and-refinement process. A Heuristic Localization Branch first generates candidate moments, while a Semantic Verification Branch evaluates their query consistency in a semantically enriched latent space. Query supervision further trains the verifier with the same candidate moment but different queries, encouraging it to distinguish the matched query from semantically similar mismatched queries. The resulting verification features are incorporated into a Progressive Refinement Loop that updates both query representations and multimodal features. To suppress potentially harmful updates, a lightweight reliability estimator gates query updates and reweights candidate contributions in feature feedback. A convergence-based adaptive stopping strategy additionally terminates unnecessary refinement during inference. Experiments on three standard MR benchmarks demonstrate the effectiveness and state-of-the-art performance of our method.
\end{abstract}


\end{frontmatter}

\section{Introduction}
\label{sec:introduction}

\begin{figure}[t]
    \centering
	\includegraphics[width=0.8\linewidth]{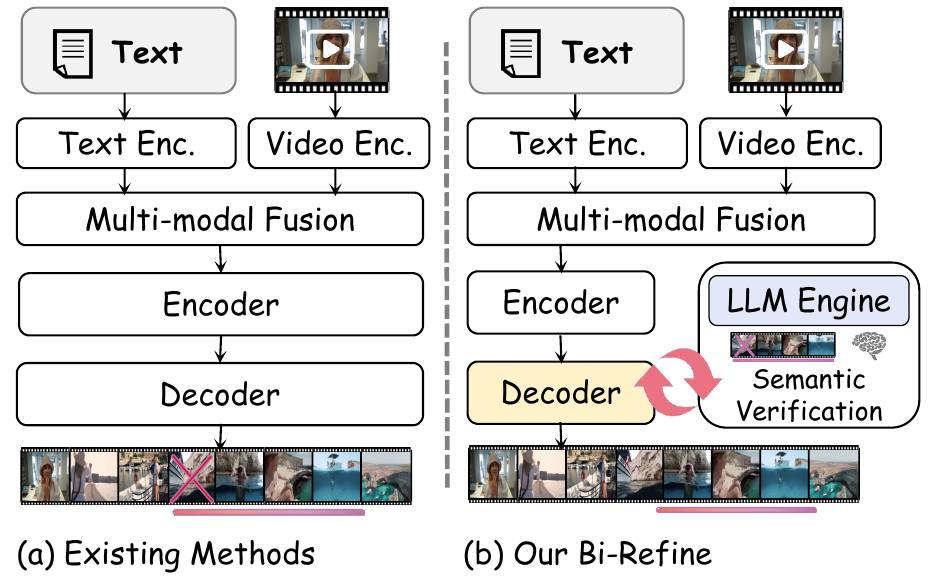}
	\caption{Motivation of Bi-Refine. 
(a) Existing retrieval frameworks mainly refine predictions within the localization process, without explicit semantic verification feedback for subsequent refinement. 
(b) Bi-Refine establishes a closed-loop semantic refinement process, where semantic consistency signals are iteratively fed back to progressively improve candidate moments.
}
	\label{fig:motivation}
\end{figure}

Moment Retrieval~(MR)~\cite{gao2017tall} aims to localize the temporal segment in an untrimmed video that semantically corresponds to a natural-language query. As a core task in video--language understanding, MR requires both query-aware semantic discrimination and accurate temporal-boundary estimation.
Existing approaches primarily optimize candidate generation, semantic estimation, and boundary prediction within a unified localization pathway. Some recent methods further employ iterative decoding to progressively update localization representations~\cite{zhao2025ld}. Nevertheless, the refinement process remains largely confined to the localization branch, without an explicit mechanism that separately evaluates candidate-level semantic consistency and feeds the resulting verification evidence back into subsequent localization, as illustrated in Figure~\ref{fig:motivation}(a).

Given the temporal redundancy and semantic ambiguity commonly present in untrimmed videos, multiple candidate segments may exhibit similar visual patterns while corresponding to different semantic meanings. Consequently, inaccurate or suboptimal predictions may persist once produced, making retrieval quality heavily dependent on the initial localization result. Moreover, iterative refinement does not necessarily improve localization monotonically: an unreliable feedback signal may move an already accurate candidate away from the target moment. These observations motivate not only explicit candidate verification, but also an assessment of whether the resulting verification evidence is sufficiently reliable for subsequent refinement.

To address the above limitation, we propose Bi-Refine, a semantic verification-guided progressive refinement framework for moment retrieval. The framework consists of three components: a Heuristic Localization Branch (HLB), a Semantic Verification Branch (SVB), and a Progressive Refinement Loop (PRL). 
Different from localization-only refinement, Bi-Refine introduces an explicit verification-and-refinement process in which candidate moments are evaluated by a separate semantic verifier and subsequently updated using semantic feedback.
Specifically, the heuristic localization branch first generates candidate temporal moments using a Transformer-based decoder. This stage focuses on producing high-recall candidate proposals that serve as the basis for subsequent refinement. The semantic verification branch then projects candidate representations into a semantically enriched latent space using a frozen LLM-based encoder and estimates their semantic consistency with the language query. To reduce the coupling between semantic relevance and temporal overlap, query supervision contrasts matched and mismatched queries while retaining identical candidate boundaries. The resulting verification signals provide complementary semantic evidence beyond the initial localization process.

To facilitate information exchange between the heuristic localization branch and the semantic verification branch, the proposed Progressive Refinement Loop (PRL) establishes bidirectional localization-verification interaction for progressive candidate refinement. PRL contains two feedback pathways: query updating and feature feedback. A lightweight reliability estimator gates query updates according to the predicted benefit of refinement and reweights candidate contributions in feature feedback, while adaptive stopping determines whether further refinement is required.

Experiments on QVHighlights~\cite{lei2021detecting}, Charades-STA~\cite{gao2017tall}, and TACoS~\cite{regneri2013grounding} evaluate the dual-branch verification-and-refinement framework. Ablations on QVHighlights further characterize the contributions and optimization behavior of query supervision, reliability-aware feedback, and adaptive stopping.

In summary, the contributions of this paper are as follows:
\begin{itemize}
\item We propose Bi-Refine, a dual-branch framework that explicitly separates temporal localization from candidate-level semantic verification, extending conventional localization-only refinement into a verification-and-refinement paradigm.

\item We introduce a Semantic Verification Branch that evaluates query-conditioned candidate consistency in a semantically enriched LLM latent space, together with query supervision that encourages discrimination between matched and mismatched queries under identical candidate evidence.

\item We design a reliability-aware Progressive Refinement Loop that feeds verification evidence back to both query representations and multimodal features and uses a lightweight reliability estimator to suppress unreliable feedback, together with adaptive stopping to avoid unnecessary iterations.

\end{itemize}

This paper is structured as follows: Section~\ref{sec:relatedWork} reviews related work in the field. In Section~\ref{sec:TripletNet}, we outline the technical details of our proposed method. Section~\ref{sec:experiments} provides the experimental results and accompanying discussions. Section~\ref{sec:discussion} provides the discussion and limitations. Finally, Section~\ref{sec:conclusion} offers a conclusion to our findings.

\section{Related Work}
\label{sec:relatedWork}

We review prior work from three perspectives: moment retrieval, semantic enhancement with vision-language models, and iterative refinement with feedback.

\subsection{Moment Retrieval}
Moment Retrieval (MR) aims to localize a temporal segment in an untrimmed video that corresponds to a natural language query. Early methods mainly follow proposal-based or proposal-free paradigms. TALL~\cite{gao2017tall} ranks sliding-window proposals, while 2D-TAN~\cite{zhang2020learning} models proposal relations on a temporal map. ExCL~\cite{ghosh2019excl} instead directly predicts temporal boundaries.
Recent DETR-style methods formulate MR as end-to-end set prediction. Moment-DETR~\cite{lei2021detecting} jointly handles moment retrieval and highlight detection, QD-DETR~\cite{moon2023query} builds query-dependent video representations, and TR-DETR~\cite{sun2024tr} exploits interactions between the two tasks. TaskWeave~\cite{yang2024task} further decouples moment retrieval and highlight detection and exchanges task-specific predictions through inter-task feedback, while LD-DETR~\cite{zhao2025ld} progressively updates localization queries using loop decoding.

Other studies improve retrieval through cross-modal modeling and task interaction. MCMN~\cite{han2023momentum} introduces momentum cross-modal contrastive learning, MHGR~\cite{wang2024modality} models multimodal relations with a heterogeneous graph, Zhou \textit{et al.}~\cite{zhou2024subtask} leverage subtask priors for joint optimization, and Ge \textit{et al.}~\cite{ge2024learning} exploit cross-instance relations for robust retrieval. Recent work in \textit{Pattern Recognition} has investigated several complementary directions. Under weak supervision, triadic temporal-semantic alignment~\cite{liu2024triadic} aligns concept, action, and event semantics, while ATSG~\cite{han2025adversarial} uses an externally pretrained semantic discriminator to guide proposal generation. For fully supervised retrieval, SMTAN~\cite{wang2025scene} constructs scene-enhanced multi-scale temporal representations, and TEA~\cite{cai2025mllm} employs an MLLM to generate temporally structured video narratives for cross-modal alignment. More recently, CG-DETR~\cite{moon2023correlation} calibrates query-dependent interaction using clip-word correlation, UPM~\cite{yang2026upm} learns unified multimodal patterns, and scale-balanced temporal grounding~\cite{zhao2026scale} explicitly addresses performance imbalance between short and long moments.

These approaches improve cross-modal representation learning, external supervision, temporal-scale modeling, or task interaction. However, semantic matching and temporal localization are generally optimized within the same retrieval pathway, leaving candidate-level semantic validity insufficiently examined after localization. Bi-Refine explores this complementary perspective by explicitly incorporating semantic verification into the refinement process.

\subsection{Semantic Enhancement with Vision-Language Models}
Vision–language pre-trained models can be broadly categorized into one-stream, dual-stream, fusion-encoder, and multimodal large language model (MLLM) architectures. ViLT~\cite{kim2021vilt} adopts a unified Transformer to jointly model visual and textual tokens, while related one-stream models~\cite{huang2021seeing} perform cross-modal interaction within a shared representation space. In contrast, dual-stream models encode the two modalities separately. CLIP~\cite{radford2021learning} aligns independently encoded visual and textual representations through contrastive learning, while RegionCLIP~\cite{zhong2022regionclip} further extends such alignment to region-level visual semantics.
Fusion-encoder models introduce intermediate cross-modal interaction to combine modality-specific representation learning with multimodal fusion. GLIPv2~\cite{zhang2022glipv2} unifies localization and vision-language understanding through grounded feature interaction, VLMo~\cite{bao2022vlmo} employs modality-specific experts with shared multimodal modeling, and FLAVA~\cite{singh2022flava} jointly learns unimodal and multimodal representations. More recently, MLLM-based approaches further leverage large language models for high-level multimodal reasoning. DeepSeek-VL2~\cite{wu2024deepseek} integrates visual representations into a language-centric architecture, Qwen2-VL~\cite{wang2024qwen2} strengthens fine-grained visual-language understanding.
Despite their strong semantic representations, pretrained semantic models are primarily exploited to enhance representation construction or proposal reasoning. Comparatively less attention has been paid to treating semantic assessment as an explicit candidate-level verification stage that participates in subsequent localization. In Bi-Refine, semantic knowledge serves as verification evidence within the refinement process rather than merely as a representation-level prior.

\subsection{Iterative Refinement and Feedback}
Iterative temporal grounding methods commonly refine decoder queries or boundaries within a single localization pathway. LD-DETR~\cite{zhao2025ld}, for example, repeatedly updates localization queries through loop decoding. TaskWeave~\cite{yang2024task} exchanges task-specific predictions between moment retrieval and highlight detection. These methods demonstrate the effectiveness of iterative prediction refinement, but their feedback is mainly derived from localization states or task-specific predictions. A remaining issue is whether the current candidate provides sufficiently trustworthy semantic evidence for guiding the next refinement step. This distinction is important because semantic relevance does not necessarily imply that applying the corresponding feedback will improve temporal localization. Bi-Refine therefore formulates iterative retrieval as a verification-guided refinement process, where candidate-level semantic assessment and feedback reliability jointly regulate subsequent localization updates.

\section{Method}
\label{sec:TripletNet}


\begin{figure*}
\centering
  \includegraphics[width=\textwidth]{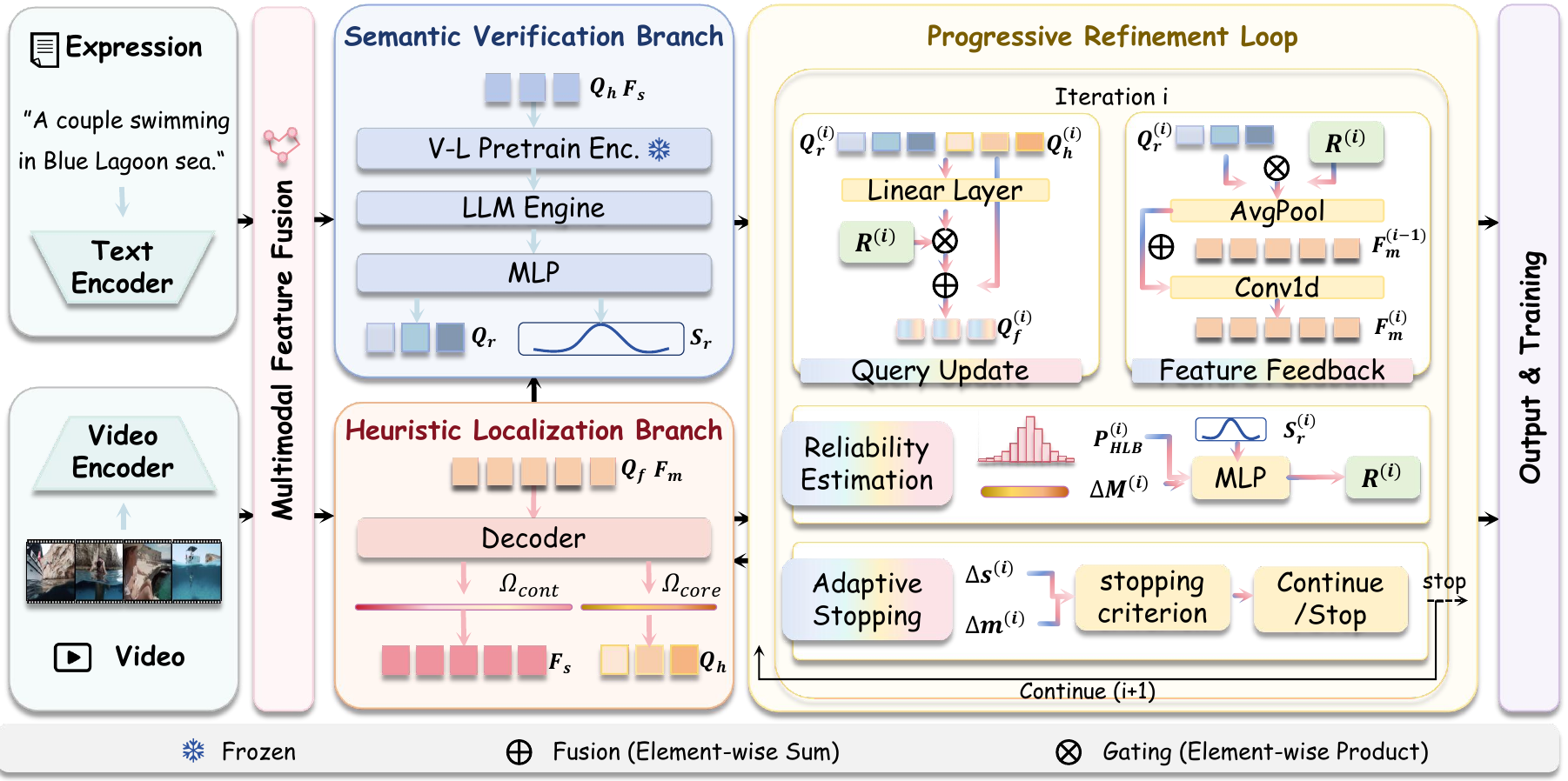}
  \caption{Overall architecture of Bi-Refine. HLB generates candidate moments and proposal features, and SVB estimates their query-conditioned semantic relevance. A reliability estimator predicts $\mathbf{R}^{(i)}$ from semantic relevance, localization confidence, and boundary variation; $R^{(i)}$ gates query feedback and weights candidate contributions in feature feedback within PRL. Query supervision is used only during training, whereas convergence-based adaptive stopping is enabled only during inference.}
  \label{fig:method}
\end{figure*}

\subsection{Overview}
Figure~\ref{fig:method} presents an overview of the Bi-Refine architecture, which consists of three key components: the Heuristic Localization Branch~(HLB), the Semantic Verification Branch~(SVB), and the Progressive Refinement Loop~(PRL). HLB generates candidate temporal moments using a DETR-style localization architecture, while SVB employs a frozen pre-trained large language model~\cite{touvron2023llama,jiang2024prior} to assess semantic consistency between candidate moments and the language query. During training, query supervision strengthens the query sensitivity of the verifier. Within PRL, semantic verification is fed back through two pathways: query updating and feature feedback. A learned reliability score gates query updating and reweights candidate contributions during feature feedback, while an adaptive stopping criterion determines whether each sample requires further refinement.

\subsubsection{Problem Formulation and Multimodal Features}
Given an untrimmed video $V=\{v_t\}_{t=1}^{L_{v}}$ consisting of $L_v$ frames and a corresponding natural language query $T=\{t_n\}_{n=1}^{L_{t}}$ with $L_t$ words, the task of Moment Retrieval (MR) is to precisely localize a temporal span $\mathbf{s} = [s_{start}, s_{end}]$ that best encapsulates the semantic intent of the query. Following the standard practice in DETR-like architectures, we represent the target moment as a normalized 2D vector $\mathbf{m} = (m_c, m_{\sigma})$, where $m_c \in [0, 1]$ and $m_{\sigma} \in [0, 1]$ denote the center time and duration of the target moment, respectively.
\newline
\textbf{Visual Features.} Following the common practice in \cite{lei2021detecting}, the input video is partitioned into a sequence of non-overlapping clips based on a fixed temporal interval. We then employ a pre-trained video encoder to extract clip-level visual representations, denoted as $\mathbf{F}_{v} = \{f_{v}^{i}\}_{i=1}^{N_v} \in \mathbb{R}^{N_v \times d_{v}}$, where $N_v$ represents the number of clips and $d_v$ indicates the feature dimensionality. To facilitate cross-modal interaction, these features are typically projected into a joint $d$-dimensional embedding space. 
\newline
\textbf{Textual Features.}  For the natural language query, we utilize a pre-trained textual backbone to generate token-level embeddings, denoted as $\mathbf{F}_t = \{f_{t}^{j}\}_{j=1}^{N_t} \in \mathbb{R}^{N_t \times d_t}$, where $N_t$ is the number of tokens and $d_t$ is the textual feature dimension. Similar to the visual branch, the textual features are mapped to the common dimension $d$ to ensure architectural consistency for subsequent fusion.

\subsection{Heuristic Localization Branch}
Existing localization models, including loop-decoding approaches such as LD-DETR~\cite{zhao2025ld}, mainly refine temporal predictions within the localization decoder. While effective, such refinement relies primarily on localization cues and lacks an explicit mechanism to assess whether the predicted moments are semantically consistent with the query. Consequently, semantically mismatched candidates may continue to be refined rather than corrected.

To address this limitation, we design the Heuristic Localization Branch (HLB) as a candidate localization module. Given multimodal features, HLB generates candidate temporal moments with an emphasis on high recall, providing a reliable basis for subsequent refinement. During the iterative refinement process, HLB continuously incorporates semantic feedback from the Semantic Verification Branch, progressively enhancing video-query alignment and localization quality across iterations.

Specifically, the video features $\mathbf{F}_v$ and the language query features $\mathbf{F}_t$ are first processed by a multimodal encoder.
This encoder employs multi-scale convolutional kernels to capture local temporal dependencies, producing an initial representation $\mathbf{F}_{m}^{(0)} \in \mathbb{R}^{L \times d}$.

Then, at each iteration $i \in \{1, \dots, N\}$, the heuristic branch acts as a “probe” to progressively identify temporal candidates from $\mathbf{F}_{m}^{(i-1)}$, formulated as,
\begin{equation} 
\{\mathbf{Q}_{h}^{(i)}, \mathbf{M}^{(i)}\} = \text{HeuristicLayer}(\mathbf{Q}_{f}^{(i-1)}, \mathbf{F}_{m}^{(i-1)}),
\end{equation}
where $\mathbf{Q}_{f}^{(0)}$ denotes the initial decoder queries, initialized as zero vectors; $\mathbf{Q}_{h}^{(i)}=\{\mathbf{Q}_{h,j}^{(i)}\}_{j=1}^{N_q}$ represents the $N_q$ queries generated by the $i$-th heuristic layer; and $\mathbf{M}^{(i)}=\{\mathbf{M}_{j}^{(i)}\}_{j=1}^{N_q}$ denotes the corresponding candidate moments. Each $\mathbf{M}_{j}^{(i)}=(c_j^{(i)},w_j^{(i)})$ encodes a normalized center and width and can equivalently be converted to start and end boundaries.

To transform temporal candidates into proposal-level features suitable for SVB, we introduce a temporal context sampling layer. For candidate $j$ at iteration $i$, its start and end boundaries define an action core region $\Omega_{\mathrm{core},j}^{(i)}$ and an extended context region $\Omega_{\mathrm{cont},j}^{(i)}$ with context ratio $\gamma=1.5$. The core region captures the candidate content, while the extended region supplies its surrounding temporal context. The memory features within the two regions are averaged separately, concatenated, and projected back to dimension $d$,
\begin{equation}
\begin{aligned}
\mathbf{F}_{s,j}^{(i)}
=
\text{MLP}\Bigg(
\Bigg[
\frac{1}{|\Omega_{\mathrm{core},j}^{(i)}|}
\sum_{t \in \Omega_{\mathrm{core},j}^{(i)}}
\mathbf{F}_{m,t}^{(i-1)};
\\
\frac{1}{|\Omega_{\mathrm{cont},j}^{(i)}|}
\sum_{t \in \Omega_{\mathrm{cont},j}^{(i)}}
\mathbf{F}_{m,t}^{(i-1)}
\Bigg]
\Bigg)
\end{aligned}
\end{equation}
where $[;]$ denotes the concatenation operation and $|\Omega|$ represents the number of temporal tokens within the region.

\subsection{Semantic Verification Branch}
To provide semantic feedback for retrieval refinement, we introduce a Semantic Verification Branch (SVB) based on a frozen large language model (LLM). Rather than directly predicting temporal boundaries, SVB evaluates whether each candidate moment semantically matches the query based on the candidates produced by the heuristic branch. By projecting candidate representations into a semantically enriched latent space, SVB produces verification signals that complement the localization process and provide additional semantic evidence for subsequent refinement.
Specifically, the process starts from the candidate representations $\mathbf{F}_{s,j}^{(i)}$ generated by the heuristic branch. These representations are mapped into the LLM latent space, where their semantic consistency with the query-conditioned candidate state is estimated. The resulting verification signals are then fed back to PRL for the next localization iteration.

To evaluate candidate moments in a semantically enriched latent space, we combine each heuristic query $\mathbf{Q}_{h,j}^{(i)}$ with its candidate representation $\mathbf{F}_{s,j}^{(i)}$. Since $\mathbf{Q}_{h,j}^{(i)}$ is produced from the previous feedback queries and multimodal memory $\mathbf{F}_m^{(i-1)}$, it encodes both query semantics and candidate-specific localization information. The fused representation is projected into the latent space of a frozen pre-trained LLM, yielding the semantic verification embedding $\mathbf{Q}_{r,j}^{(i)}$.
To bridge the dimensional gap between the heuristic space ($d$) and the LLM latent space ($d_{llm}$), we introduce two nonlinear projection functions, $\phi_{in}(\cdot)$ and $\phi_{out}(\cdot)$, each implemented as a two-layer multilayer perceptron followed by layer normalization. These projectors facilitate cross-space embedding alignment and stabilize feature representations across the two spaces. Formally, the semantic projection process is defined as,

\begin{equation}
\label{LLaMA}
    \mathbf{Q}_{r,j}^{(i)} = \phi_{out}\big(\text{LLM}(\phi_{in}(\mathbf{Q}_{h,j}^{(i)} + \mathbf{F}_{s,j}^{(i)}))\big).
\end{equation}

Finally, a linear classification head derives the candidate relevance score $\mathbf{S}_{r,j}^{(i)}$,

\begin{equation}
\mathbf{S}_{r,j}^{(i)} = \sigma\big(\text{MLP}(\mathbf{Q}_{r,j}^{(i)})\big),
\end{equation}
where $\sigma(\cdot)$ denotes the sigmoid activation. The relevance score $S_{r,j}^{(i)}$ measures the semantic consistency of candidate $j$ with the query-conditioned retrieval context.

We further conduct semantic encoder replacement studies in SVB by substituting the frozen LLaMA encoder with lightweight MLP and Transformer alternatives. The results demonstrate that the effectiveness of SVB mainly stems from semantic consistency estimation, while richer semantic representations provided by LLaMA further improve the quality of verification signals.

\subsection{Progressive Refinement Loop}
The Progressive Refinement Loop (PRL) enables bidirectional interaction between
HLB and SVB by feeding candidate-level verification evidence back to subsequent
localization stages. PRL contains two feedback pathways, namely query updating and feature feedback. Reliability estimation gates the query update and reweights candidate contributions in feature feedback, while adaptive stopping determines whether further refinement is required.


\subsubsection{Bidirectional Feedback}
\noindent\textbf{Query Updating.}
In the initial stage, shallow multimodal features often struggle to achieve fine-grained understanding of textual semantics, which can lead to semantic drift in the search queries and hinder the model’s focus on visual content that matches the query. To address this issue, we incorporate verification signals from the Semantic Verification Branch (SVB) into the query representations, enabling the model to progressively refine candidate moments and improve video-query alignment.

Specifically, we employ the predicted reliability $\mathbf{R}^{(i)}$ as a gate to regulate the query updating process. The updated query $\mathbf{Q}_f^{(i)}$ is obtained by fusing the heuristic query $\mathbf{Q}_h^{(i)}$ and the semantic verification query $\mathbf{Q}_r^{(i)}$, while feedback that is unlikely to improve localization is attenuated.
The process is described as follows,
 \begin{equation}
 \mathbf{Q}_f^{(i)}
 =
 \mathbf{Q}_h^{(i)}
 +
 \mathbf{R}^{(i)}\odot
 \phi_{L1}\!\left([\mathbf{Q}_h^{(i)};\mathbf{Q}_r^{(i)}]\right),
 \label{eq:reliable_query_update}
\end{equation}
where $\phi_{L1}$ denotes a linear projection of the concatenated queries and $\odot$ denotes element-wise multiplication.

\noindent\textbf{Feature Feedback.}
To facilitate bidirectional interaction between the Semantic Verification Branch (SVB) and the Heuristic Localization Branch (HLB), we introduce a feature feedback mechanism that incorporates verification signals into the multimodal feature space. Specifically, at the $i$-th iteration, we perform a reliability-aware aggregation of the semantic verification query $\mathbf{Q}_{r}^{(i)}$ to obtain a global verification representation $\mathbf{f}_\mathrm{global}^{(i)}$,


\begin{equation}
\mathbf{f}_{\mathrm{global}}^{(i)}
=
\frac{
\sum_{j=1}^{N_q}
\mathbf{R}_{j}^{(i)}\mathbf{Q}_{r,j}^{(i)}
}{
\sum_{j=1}^{N_q}\mathbf{R}_{j}^{(i)}+\epsilon
}
\in \mathbb{R}^{1\times d}.
\label{eq:reliable_feature_feedback}
\end{equation}

This normalized representation aggregates verification signals from the candidate proposals while preventing low-reliability candidates from dominating the global feedback prior.

The aggregated feedback is broadcast along the temporal dimension and aligned with the feature memory $\mathbf{F}_{m}^{(i-1)} \in \mathbb{R}^{L \times d}$ to form a verification-aware prior. The prior is then concatenated with $\mathbf{F}_{m}^{(i-1)}$ to construct an enhanced feature representation. To integrate the feedback effectively while preserving local temporal dependencies, the concatenated features are further processed using a one-dimensional temporal convolution, producing the final feedback features,
\begin{equation}
    \mathbf{F}_{m}^{(i)} = \text{Conv1D}([\mathbf{F}_{m}^{(i-1)};\mathbf{f}_\mathrm{global}^{(i)}]).
\end{equation}
In this process, the convolution operation integrates global semantic guidance with local temporal features.
The resulting features $\mathbf{F}_{m}^{(i)}$ are then fed back into the heuristic branch, enabling it to refine temporal predictions with enhanced semantic awareness.

\subsubsection{Refinement Control}
\noindent\textbf{Reliability Estimation.}
A high relevance score does not necessarily imply that applying the corresponding feedback will improve temporal localization. We therefore introduce a lightweight reliability estimator that predicts whether the current refinement direction is beneficial. For the $j$-th candidate at iteration $i$, its input is
\begin{equation}
\mathbf{z}_{j}^{(i)}
=
\left[
\mathbf{S}_{r,j}^{(i)};
\mathbf{P}_{\mathrm{HLB},j}^{(i)};
\left|c_{j}^{(i)}-c_{j}^{(i-1)}\right|;
\left|w_{j}^{(i)}-w_{j}^{(i-1)}\right|
\right],
\label{eq:reliability_input}
\end{equation}
where $\mathbf{P}_{\mathrm{HLB},j}^{(i)}$ is the foreground softmax confidence predicted by the localization classification head, and $c_j^{(i)}$ and $w_j^{(i)}$ denote the center and width of the candidate. At the first iteration, where no preceding prediction is available, the two variation terms are set to zero. The four-dimensional input is processed by a two-layer MLP with a hidden dimension of 64,
\begin{equation}
\mathbf{R}_{j}^{(i)}
=
\sigma\!\left(
\mathbf{W}_{2}\operatorname{ReLU}
\left(\mathbf{W}_{1}\mathbf{z}_{j}^{(i)}+\mathbf{b}_{1}\right)
+\mathbf{b}_{2}
\right),
\label{eq:feedback_reliability}
\end{equation}
where $\mathbf{R}_{j}^{(i)}\in[0,1]$ denotes the predicted feedback reliability.

\noindent\textbf{Adaptive Stopping.}
For each sample, once two successive states are available, we compute the mean change in relevance scores and candidate boundaries,
\begin{equation}
\Delta_{s}^{(i)}
=
\frac{1}{N_q}
\sum_{j=1}^{N_q}
\left|
\mathbf{S}_{r,j}^{(i)}-\mathbf{S}_{r,j}^{(i-1)}
\right|,
\end{equation}
\begin{equation}
\Delta_{m}^{(i)}
=
\frac{1}{2N_q}
\sum_{j=1}^{N_q}
\left\|
\mathbf{M}_{j}^{(i)}-\mathbf{M}_{j}^{(i-1)}
\right\|_{1}.
\end{equation}
The sample stops further refinement when
\begin{equation}
\Delta_{s}^{(i)}<0.01
\quad\land\quad
\Delta_{m}^{(i)}<0.02.
\label{eq:adaptive_stopping}
\end{equation}
Once a sample satisfies the stopping criterion, its query, candidate boundaries, and video memory are frozen, and its current state is retained as the final active state. 
The loop terminates immediately when all samples in the mini-batch have stopped. The prediction and ranking heads operate on each sample's final active state to select the output moment $\hat{\mathbf{m}}=(\hat{m}_c,\hat{m}_{\sigma})$; adaptive stopping changes only the number of state updates and does not perform selection among iterations.

\subsection{Training Objective}
The training objective of Bi-Refine jointly optimizes localization, semantic verification, and feedback reliability. The heuristic branch is trained using a DETR-style objective, while the verification branch is supervised by candidate relevance and query consistency. The reliability estimator is trained to identify updates that improve temporal overlap.

\subsubsection{Localization Supervision}

For temporal localization, we adopt the training paradigm of LD-DETR~\cite{zhao2025ld}. The DETR-based head predicts the center and duration of temporal moments together with foreground confidence and localization quality. The moment-retrieval loss $\mathcal{L}_{detr}$ consists of Smooth L1 boundary regression, Generalized IoU regression, foreground classification, and localization-quality prediction,

\begin{equation}
\label{eq:localization_loss}
\begin{split}
\mathcal{L}_{detr} =\ & \lambda_{L_{1}}\mathcal{L}_{1}(\mathbf{m}^{*}, \hat{\mathbf{m}})
+ \lambda_{GIoU}\mathcal{L}_{GIoU}(\mathbf{m}^{*}, \hat{\mathbf{m}}) \\
& + \lambda_{cls}\mathcal{L}_{CE}(y, \hat{y})
+ \lambda_{qual}\mathcal{L}_{CE}(IoU, \widehat{IoU}),
\end{split}
\end{equation}
where $\mathbf{m}^{*}$ and $\hat{\mathbf{m}}$ denote the ground-truth and predicted moment center and duration, respectively. The variables $y$ and $IoU$ represent the foreground label and localization-quality target, respectively. The four loss weights are distinguished because boundary overlap and quality prediction supervise different outputs. Following LD-DETR~\cite{zhao2025ld}, we further retain its original
highlight detection head and saliency loss $\mathcal{L}_{\mathrm{sal}}$.

\subsubsection{Semantic Verification Supervision}
To supervise the semantic verification branch, we introduce a
relevance loss $\mathcal{L}_{\mathrm{rel}}$.
The loss encourages the relevance score $S_r$ to distinguish
ground-truth-matched candidates from unmatched candidates under the
query-conditioned retrieval context.
Given the optimal bipartite matching obtained via the Hungarian algorithm, we define,
\begin{equation}
\mathcal{L}_{rel} = -\frac{1}{N_q} \sum_{j=1}^{N_q} [y_j \log(\mathbf{S}_{r,j}) + (1-y_j) \log(1-\mathbf{S}_{r,j})],
\end{equation}
where $y_j=1$ indicates that the $j$-th candidate is matched to a
ground-truth moment by the Hungarian algorithm, and $y_j=0$ otherwise.

\subsubsection{Query Supervision}
The relevance labels derived from Hungarian matching may partially reflect
temporal localization quality, making it difficult to determine whether the
verifier truly captures query--candidate semantic compatibility. To provide
more explicit semantic supervision, we construct query pairs
while keeping the candidate evidence fixed.

For each query, we first obtain a global textual representation by masked
mean pooling over its valid token features, followed by $\ell_2$
normalization. Given the normalized query representation
$\bar{\mathbf{t}}_b$ of sample $b$, we select a semantic hard negative from
the other videos in the same mini-batch as
\begin{equation}
k^{*}
=
\arg\max_{k:\,k\neq b,\;v_k\neq v_b}
\bar{\mathbf{t}}_{b}^{\top}\bar{\mathbf{t}}_{k},
\label{eq:hard_negative}
\end{equation}
where $v_b$ denotes the video associated with query $b$. The selected query
$\mathbf{T}^{-}=\mathbf{T}_{k^{*}}$ is semantically similar to the positive
query $\mathbf{T}^{+}$ but originates from a different video.

For candidate $j$, we keep its temporal boundary $\mathbf{M}_{j}^{(i)}$
and the corresponding sampled evidence $\mathbf{F}_{s,j}^{(i)}$ fixed.
The selected negative text $\mathbf{T}^{-}$ is only used to recompute the
query-conditioned heuristic representation, denoted as
$\mathbf{Q}_{h,j}^{(i),-}$, while the candidate evidence from the positive
localization path is reused.
Let $g_{\mathrm{SVB}}(\cdot,\cdot)$ denote the semantic relevance scoring
process defined in Eqs.~(3)--(4). The positive and negative relevance scores
are computed as,
\begin{equation}
S_{r,j}^{(i),+}
=
g_{\mathrm{SVB}}
\left(
\mathbf{Q}_{h,j}^{(i),+},
\mathbf{F}_{s,j}^{(i)}
\right),
\qquad
S_{r,j}^{(i),-}
=
g_{\mathrm{SVB}}
\left(
\mathbf{Q}_{h,j}^{(i),-},
\mathbf{F}_{s,j}^{(i)}
\right),
\label{eq:counterfactual_scores}
\end{equation}
where $\mathbf{Q}_{h,j}^{(i),+}$ and
$\mathbf{Q}_{h,j}^{(i),-}$ denote the query-conditioned heuristic
representations obtained with the matched query $\mathbf{T}^{+}$ and the
selected negative query $\mathbf{T}^{-}$, respectively. Importantly, the
candidate boundary $\mathbf{M}_{j}^{(i)}$ and its sampled evidence
$\mathbf{F}_{s,j}^{(i)}$ are shared by the two paths. Therefore, the score
difference mainly reflects the compatibility between the textual condition
and the same candidate evidence rather than differences in localization
quality. We then employ the margin-based ranking loss as,
\begin{equation}
\mathcal{L}_{cf}
=
\frac{1}{N_{cf}}
\sum_{j=1}^{N_{cf}}
\max\left(0,m_{cf}-S_{r,j}^{+}+S_{r,j}^{-}\right),
\label{eq:counterfactual_loss}
\end{equation}
where $m_{cf}$ is the ranking margin and $N_{cf}$ is the number of valid pairs. Samples without a valid cross-video negative are excluded from $N_{cf}$. Since the candidate remains fixed, this loss encourages the score difference to reflect textual compatibility.

\subsubsection{Feedback Reliability Supervision}
After obtaining the proposals of iteration $i+1$, the reliability estimator is supervised by whether the corresponding refinement improves temporal overlap,
\begin{equation}
y_{j}^{(i)}
=
\mathbb{I}\!\left[
\operatorname{IoU}\!\left(\mathbf{M}_{j}^{(i+1)},\mathbf{M}_{j}^{*}\right)
>
\operatorname{IoU}\!\left(\mathbf{M}_{j}^{(i)},\mathbf{M}_{j}^{*}\right)
+\delta
\right],
\label{eq:improvement_label}
\end{equation}
where $\mathbf{M}_{j}^{*}$ is the matched ground-truth moment and $\delta=0.02$. Both $\mathbf{M}_{j}^{(i)}$ and $\mathbf{M}_{j}^{(i+1)}$ are detached when constructing this stop-gradient target. The reliability loss is computed only for candidates assigned to a ground-truth moment, and the same matched target is used to compare two successive states. Consequently, the IoU comparison supervises the reliability estimator without propagating gradients through candidate boundaries. We use a binary cross-entropy loss,
\begin{equation}
\mathcal{L}_{imp}
=
-\frac{1}{N_{imp}}
\sum_{i,j}
\left[
y_{j}^{(i)}\log\mathbf{R}_{j}^{(i)}
+
\left(1-y_{j}^{(i)}\right)
\log\left(1-\mathbf{R}_{j}^{(i)}\right)
\right].
\label{eq:improvement_loss}
\end{equation}
Here, \(y_j^{(i)}\) indicates whether the \(j\)-th candidate obtains a meaningful localization improvement after the \(i\)-th refinement. Specifically, \(y_j^{(i)}=1\) if the IoU with the matched ground-truth moment increases by more than \(\delta\), and \(y_j^{(i)}=0\) otherwise. Accordingly, \(\mathbf{R}_j^{(i)}\) is trained to estimate whether another feedback update is likely to improve the current candidate. The candidate boundaries used to construct \(y_j^{(i)}\) are detached, such that \(\mathcal{L}_{imp}\) only supervises the reliability estimator without affecting the localization branch.


\subsubsection{Final Objective}
The overall training objective is defined as,

\begin{equation}
\mathcal{L}_{overall}
=
\mathcal{L}_{detr}
+
\mathcal{L}_{sal}
+
\lambda_{rel}\mathcal{L}_{rel}
+
\lambda_{cf}\mathcal{L}_{cf}
+
\lambda_{imp}\mathcal{L}_{imp},
\label{eq:final_objective}
\end{equation}
where $\lambda_{rel}$, $\lambda_{cf}$, and $\lambda_{imp}$ balance relevance, query supervision, and reliability supervision, respectively. 

\section{Experiments}
\label{sec:experiments}
We conduct extensive experiments and compare with prior studies on QVHighlights~\cite{lei2021detecting}, Charades-STA~\cite{gao2017tall} and TACoS~\cite{regneri2013grounding} datasets. We also perform comprehensive ablation studies to validate our key contributions.

\begin{table*}[tb]
  \caption{Comparison on QVHighlights. $^\dagger$ indicates the use of an LLM.}
  \label{tab:qvhighlights_main}
  \centering
  \resizebox{\linewidth}{!}{
  \begin{tabular}{l c c c c c c c}
    \toprule
    \multirow{3}{*}{Method} & \multicolumn{5}{c}{MR} & \multicolumn{2}{c}{HD} \\  
    \cmidrule(lr){2-6}
    \cmidrule(lr){7-8}
    & \multicolumn{2}{c}{R1} & \multicolumn{3}{c}{mAP} & \multicolumn{2}{c}{$\geq$Very Good} \\ 
    \cmidrule(lr){2-3}
    \cmidrule(lr){4-6}
    \cmidrule(lr){7-8}
    & $@$0.5 & $@$0.7 & $@$0.5  & $@$0.75 & Avg. & mAP & HIT$@$1 \\
    \midrule
    M-DETR~\cite{lei2021detecting} (NeurIPS'2021) & 52.89 & 33.02 & 54.84 & 29.40 & 30.73 & 35.69 & 55.60 \\
    QD-DETR~\cite{moon2023query} (CVPR'2023) & 62.52 & 44.98 & 62.62 & 39.88 & 39.86 & 38.64 & 62.40 \\
    UniVTG~\cite{lin2023univtg} (ICCV'2023) & 66.65 & 52.19 & 64.37 & 46.68 & 45.18 & 40.18 & 64.77  \\
    TaskWeave~\cite{yang2024task} (CVPR'2024) & 64.26 & 50.06 & 65.39 & 46.47 & 45.38 & - & - \\
    TR-DETR~\cite{sun2024tr} (AAAI'2024) & 67.10 & {51.48} & 66.27 & 45.09 & 46.42 & 40.55 & 64.77 \\
    LLMEPET$^\dagger$~\cite{jiang2024prior} (ACM MM'2024) & 66.58 & 51.10 & - & - & 46.24 & - & - \\

 AMQP~\cite{zeng2024improving} (TCSVT'2025) & 67.81 & - &- &- &- &-&-\\
LD-DETR~\cite{zhao2025ld} (arXiv'2025) &\underline{67.87}& 51.35 &67.23& 47.29 & 47.18 & \underline{41.39} &66.39\\
MPDC~\cite{hu2026mismatched} (TCSVT'2026) & 66.61 & 49.10 & 65.69 & 43.28 & 43.96 & 40.64 & 66.49\\
CG-DETR~\cite{moon2023correlation} (PR'2026) & 67.35 & 52.06 & 65.57 & 45.73 & 44.93 & 40.79 & \underline{66.71} \\
UPM~\cite{yang2026upm} (PR'2026) & 67.42 & \underline{52.45} & \underline{67.74} & \underline{48.60} & \underline{47.52} & 40.76 & 66.52 \\
    Bi-Refine (Ours) & \textbf{70.45} & \textbf{54.00} & \textbf{68.82} & \textbf{48.82} & \textbf{48.00} & \textbf{42.43} & \textbf{69.94} \\
    \bottomrule
  \end{tabular}
  }
\end{table*}

\begin{table*}[tb]
  \caption{Comparison on TACoS and Charades-STA.}
\label{tab:cross_dataset}
  \centering
  \resizebox{1\linewidth}{!}{
  \begin{tabular}{l c c c c c c c c}
    \toprule
   \multirow{2}{*}{Method} & \multicolumn{4}{c}{TACoS} & \multicolumn{4}{c}{Charades-STA} \\  
    \cmidrule(r){2-5}
    \cmidrule(r){6-9}
 & R$@$0.3 & R$@$0.5 & R$@$0.7 & mIoU &  R$@$0.3 & R$@$0.5 & R$@$0.7 & mIoU\\
\midrule
M-DETR~\cite{lei2021detecting} (NeurIPS'2021) & 38.0 & 24.7 & 12.0 & 25.5 & 65.8 & 52.1 & 30.6 & 45.5  \\
MomentDiff~\cite{li2023momentdiff} (NeurIPS'2023) & 44.8 & 33.7 & - & - & - & 55.6 & 32.4 & -  \\
UniVTG~\cite{lin2023univtg} (ICCV'2023) & - & - & - & - & 68.0 & 57.2 & 36.1 & - \\
TR-DETR~\cite{sun2024tr} (AAAI'2024) & - & - & - & - & - & 57.6 & 33.5 & -  \\
LLMEPET$^\dagger$~\cite{jiang2024prior} (ACM MM'2024) & 52.7 & - & 22.8 & 36.6 & 70.9 & - & 36.5 & 50.3 \\
 FMRN~\cite{zhao2024fine} (TCSVT'2025)  & 52.2 & 39.8 & - & -&73.4 & \underline{62.8} & \underline{41.9} &-\\
LD-DETR~\cite{zhao2025ld} (arXiv'2025) & \underline{57.6} & \underline{44.3} & \underline{26.2} & \underline{40.3} & \underline{73.9} & {62.6} & {41.6} & \underline{53.4} \\
CG-DETR~\cite{moon2023correlation} (PR'2026) & 54.4 & 39.5 & 23.4 & 37.4 & 70.4 & 58.4 & 36.3 & 50.1 \\
UPM~\cite{yang2026upm} (PR'2026) & 51.2 & 38.8 & 24.0 & 36.6 & 70.3 & 59.2 & 38.1 & 50.6 \\
Bi-Refine (Ours) &  \textbf{58.2} & \textbf{45.9} & \textbf{27.1} & \textbf{41.5} & \textbf{75.1} & \textbf{63.6} & \textbf{42.4} &  \textbf{54.3}  \\

    \bottomrule
  \end{tabular}
  }
\end{table*}

\subsection{Datasets and Evaluation Metrics}
\noindent\textbf{QVHighlights}~\cite{lei2021detecting} dataset comprises 10,148 videos, 18,367 annotated moments, and 10,310 sentences. Each annotation is accompanied by a saliency score on a five-level ordinal scale (from \enquote{very poor} to \enquote{excellent}), enabling simultaneous video grounding and highlight detection. In our experiments, we report results on the official validation set, following the standard protocol.
\newline
\textbf{Charades-STA}~\cite{gao2017tall} is constructed from the Charades video collection and contains 16,128 moment-query pairs over 6,672 videos. The standard split contains 12,408 training pairs from 5,338 videos and 3,720 test pairs from 1,334 videos. The source videos are approximately 30 seconds long on average, whereas the annotated moments last about 8.1 seconds.
\newline
\textbf{TACoS}~\cite{regneri2013grounding} contains 127 long cooking videos and 18,818 moment-query pairs. Following the standard temporal-grounding split, we use 10,146 pairs for training, 4,589 for validation, and 4,083 for testing. The videos range from 1 to 23 minutes and therefore require localization over substantially longer temporal contexts than Charades-STA.

\noindent\textbf{Evaluation metrics.} We adopt the standard evaluation metrics from TR-DETR~\cite{sun2024tr} and TD-DETR~\cite{zhou2025devil}. For QVHighlights, we report Recall@1 (R1) at IoU thresholds of 0.5 and 0.7, mAP at IoU thresholds of 0.5 and 0.75, and the average mAP over 10 IoU thresholds from 0.5 to 0.95~\cite{lei2021detecting}. For highlight detection on QVHighlights, we additionally report mAP and HIT@1 under the ``Very Good'' criterion. For TACoS and Charades-STA, we evaluate Recall@1 at IoU $\in \left\{0.3,0.5,0.7\right\}$ and mean IoU (mIoU). Unless explicitly stated otherwise, QVHighlights results and all ablations use the official validation split, whereas Charades-STA and TACoS results use their standard test splits.
\subsection{Implementation Details}
For QVHighlights, visual features are extracted from SlowFast~\cite{feichtenhofer2019slowfast} and CLIP~\cite{radford2021learning}, while textual features are obtained via CLIP’s text encoder~\cite{radford2021learning}. The model is trained for 200 epochs with a batch size of 32 and a learning rate of 1e-4.
For Charades-STA, visual features follow TR-DETR~\cite{sun2024tr} and are extracted using VGG~\cite{simonyan2015very}, I3D~\cite{carreira2017quo}, SlowFast~\cite{feichtenhofer2019slowfast}, and CLIP~\cite{radford2021learning}, with textual features from GloVe~\cite{pennington2014glove}. Training runs for 100 epochs with a batch size of 8 and a learning rate of 1e-4.
For TACoS, visual features are extracted using C3D~\cite{tran2015learning}, and textual features are derived from GloVe~\cite{pennington2014glove}. The model is trained for 200 epochs with a batch size of 32 and a learning rate of 2$\times$1e-4. 

For all datasets, the hidden dimension is $d=256$, the number of localization queries is $N_q=10$, the maximum number of refinement iterations is $N=3$, and the temporal context ratio is $\gamma=1.5$. The Semantic Verification Branch employs a frozen LLaMA-7B model~\cite{touvron2023llama}; only the surrounding retrieval network and the lightweight input/output projectors are optimized. Candidate representations are mapped to the LLM latent dimension $d_{llm}=4096$, and features from the 14th-17th Transformer layers are used for semantic consistency estimation following LLMEPET~\cite{jiang2024prior}. The reliability estimator has a $4$-$64$-$1$ MLP architecture. We set $\lambda_{rel}=1$, $\lambda_{cf}=0.5$, $\lambda_{imp}=0.5$, and $\delta=0.02$. During training, all samples execute three refinement iterations. During inference, stopping is evaluated sample-wise with $\tau_s=0.01$ and $\tau_m=0.02$, as defined in Eq.~\eqref{eq:adaptive_stopping}. All experiments are conducted on a single NVIDIA RTX 3090 GPU.

\subsection{Comparison with State-of-the-Art Methods}
Table~\ref{tab:qvhighlights_main} presents the performance of the Bi-Refine on QVHighlights. It improves R1@0.5 and R1@0.7 by 2.58 and 2.65 percentage points, respectively, over LD-DETR, demonstrating the high-recall capability of the Progressive Refinement Loop (PRL). It also obtains the best highlight-detection results among the methods included in the table. These gains suggest that candidate-level semantic verification and iterative feedback improve the identification of query-relevant moments.
Our model is built upon LD-DETR~\cite{zhao2025ld} as the baseline and achieves notable performance improvements over it, validating the effectiveness of the proposed approach.
Furthermore, Table~\ref{tab:cross_dataset} reports the core-model results on TACoS and Charades-STA. The improvements over the localization baseline on both datasets indicate that the dual-branch verification-and-refinement design generalizes beyond QVHighlights.

\begin{figure*}[t]
\centering
  \includegraphics[width=\textwidth]{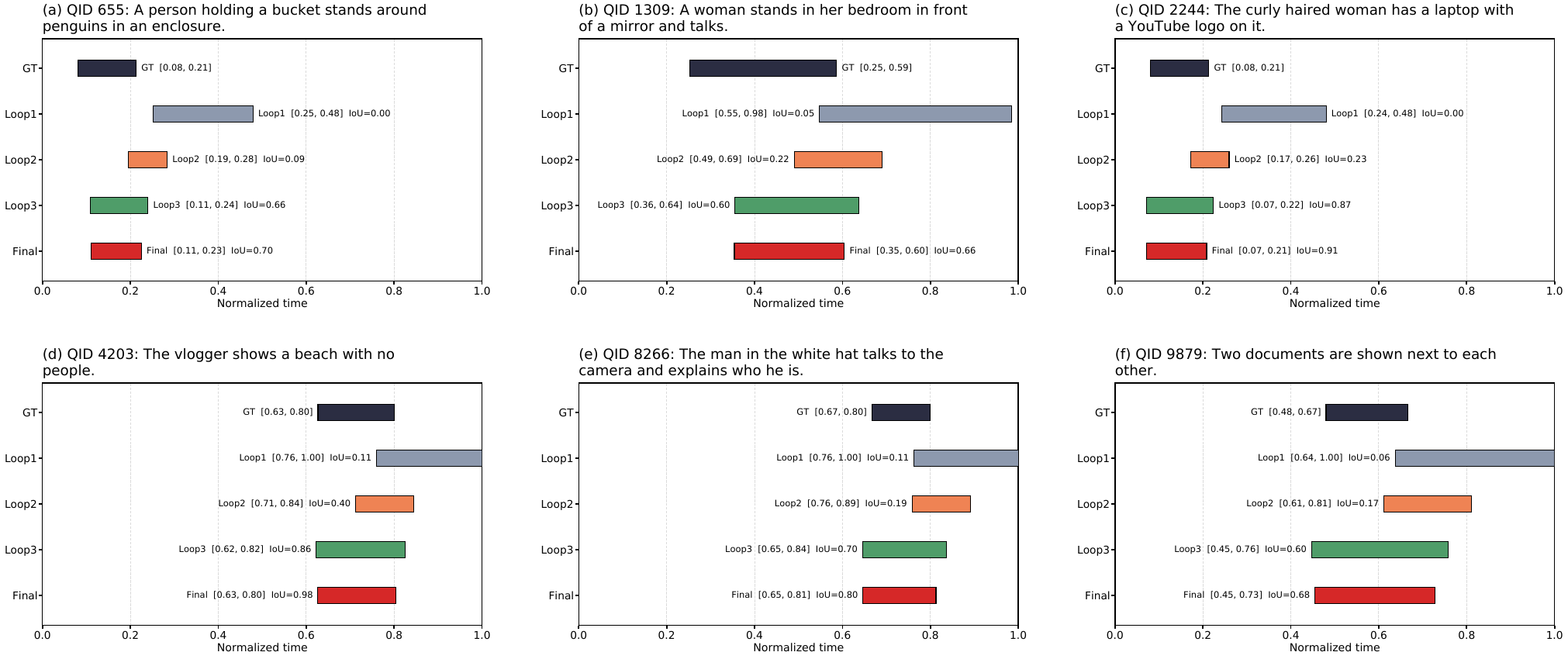}
  \caption{Qualitative visualization of PRL. Loop 1-Loop 3 show the selected intermediate candidate at each refinement iteration, while Final denotes the output selected by the prediction and ranking heads from the final active state.}
  \label{fig:qualitative_loop}
\end{figure*}

\begin{table}[t]
  \caption{Ablation study of the core branches on QVHighlights.}
  \centering
  \small
  \label{tab:core_components}
  \begin{tabular}{l|ccccc}
        \toprule
        Module & R1@0.5 & R1@0.7 & mAP@0.5 & mAP@0.75 & Avg. \\
        \hline
         HLB    &67.87& 51.35 &67.23&47.29&47.18 \\
         SVB   &68.52&51.87 &66.92&47.43& 46.32\\
         Bi-Refine & 70.45 & 54.00 &68.82 & 48.82 & 48.00 \\
        \bottomrule
      \end{tabular}
\end{table}


\begin{table}[t]
  \centering
  \small
  \caption{Ablation study of semantic encoders across QVHighlights, TACoS, and Charades-STA using the core Bi-Refine configuration.}
  \label{tab:semantic_encoder_ablation}
  \begin{tabular}{llcccc}
  \toprule
  Dataset & Encoder & R$@$0.5 & R$@$0.7 & Avg. & mIoU \\
  \midrule

  \multirow{3}{*}{QVHighlights}
  & MLP         & 68.58 & 52.00 & 47.09 & - \\
  & Transformer & 68.97 & 52.39 & 47.35 & - \\
  & LLaMA-7B    & \textbf{70.45} & \textbf{54.00}
                & \textbf{48.00} & - \\

  \midrule

  \multirow{3}{*}{TACoS}
  & MLP         & 41.5 & 24.8 & - & 38.6 \\
  & Transformer & 43.1 & 25.3 & - & 39.7 \\
  & LLaMA-7B    & \textbf{45.9} & \textbf{27.1}
                & - & \textbf{41.5} \\

  \midrule

  \multirow{3}{*}{Charades-STA}
  & MLP         & 59.8 & 39.7 & - & 51.4 \\
  & Transformer & 58.6 & 39.3 & - & 50.8 \\
  & LLaMA-7B    & \textbf{63.6} & \textbf{42.4}
                & - & \textbf{54.3} \\

  \bottomrule
  \end{tabular}
  \end{table}

\begin{table}[t]
  \caption{Ablation study of feature-fusion strategies in SVB using the core Bi-Refine configuration.}
  \label{tab:feature_fusion}
  \small
  \centering
  \begin{tabular}{l|ccccc}
        \toprule
        Strategy & R1@0.5 & R1@0.7 & mAP@0.5 & mAP@0.75 & Avg. \\
        \hline
         concatenation   &69.42& 52.71 &68.72&48.09&47.77 \\
         addition   & 70.45 & 54.00 &68.82 & 48.82 & 48.00 \\
        \bottomrule
      \end{tabular}
\end{table}


\begin{table}[t]
  \caption{Ablation of query supervision (QS), reliability-aware feedback (RF), and adaptive stopping (AS).}
  \label{tab:reliable_feedback_ablation}
  \centering
  \small
  \begin{tabular}{ccc|ccccc}
        \toprule
        QS & RF & AS & R1@0.5 & R1@0.7 & mAP@0.5 & mAP@0.75 & Avg. \\
        \midrule
         &  &  & 67.87 & 51.35 & 67.23 & 47.29 & 47.18 \\
        \checkmark &  &  & 68.45 & 52.26 & 67.54 & 47.91 & 47.47 \\
         & \checkmark &  & 68.00 & 52.97 & 67.64 & 48.00 & 47.23 \\
         & \checkmark & \checkmark & 69.81 & 53.94 & 68.44 & 48.82 & 48.00 \\
        \checkmark & \checkmark & \checkmark &
        \textbf{70.45} & \textbf{54.00} & \textbf{68.82} &
        \textbf{48.82} & \textbf{48.00} \\
        \bottomrule
  \end{tabular}
\end{table}

\subsection{Ablation Studies and Analysis}
\subsubsection{Core Components}
To evaluate bidirectional feedback, we ablate the core components of Bi-Refine in Table~\ref{tab:core_components}. HLB only provides the base temporal-localization capability without semantic verification feedback. SVB only follows the semantic-modeling paradigm of LLMEPET~\cite{jiang2024prior}, where multimodal representations are
processed by a pretrained LLaMA model.
Although semantic modeling improves several recall metrics, SVB alone
does not consistently improve overall localization performance,
highlighting the importance of fine-grained temporal modeling. Bi-Refine: Combining HLB and SVB through PRL achieves the strongest overall performance in this branch ablation, supporting the use of verification feedback for temporal localization.

\subsubsection{Semantic Verification Encoder}
Table~\ref{tab:semantic_encoder_ablation} compares different semantic
  encoders in the Semantic Verification Branch. Overall, LLaMA-7B
  consistently outperforms both the MLP and Transformer encoders across
  all three datasets. On QVHighlights, LLaMA-7B improves R$@$0.5,
  R$@$0.7, and average mAP over the Transformer by 1.48, 1.61, and 0.65
  points, respectively. On TACoS, it surpasses the Transformer by 2.8
  points in R$@$0.5, 1.8 points in R$@$0.7, and 1.8 points in mIoU.
  Similar improvements are observed on Charades-STA, where LLaMA-7B
  yields gains of 5.0, 3.1, and 3.5 points in R$@$0.5, R$@$0.7, and mIoU,
  respectively. These consistent gains demonstrate that the richer
  semantic representations provided by LLaMA-7B enable more effective
  semantic verification and more accurate temporal localization than
  lightweight MLP and Transformer encoders.

\begin{table}[t]
    \centering
    \caption{Inference efficiency.}
    \small
  \label{tab:inference_efficiency}
\begin{tabular}{lcccc}
\toprule
 & Baseline & Loop 1 & Loop 2 & Loop 3\\ \hline
Infer time (ms) &   26.56  &29.79&  30.08  &  30.64  \\
\bottomrule
\end{tabular}
\end{table}

\begin{figure*}[t]
\centering
  \includegraphics[width=\textwidth]{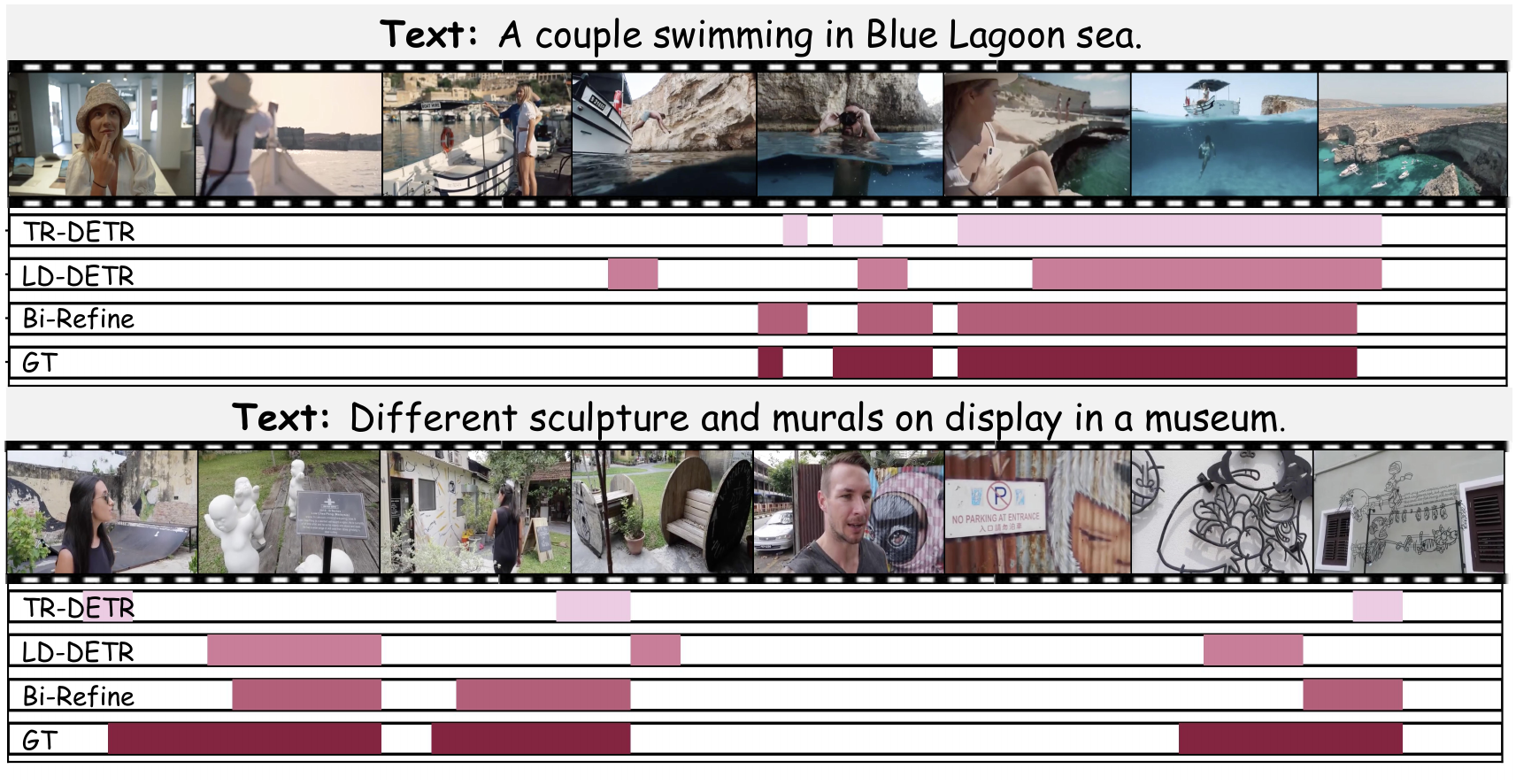}
\caption{Qualitative results of Bi-Refine on the QVHighlights validation split.}
  \label{fig:qualitative}
\end{figure*}

\subsubsection{Feature Fusion}
To evaluate the effectiveness of different feature fusion strategies in the semantic verification branch, we compare addition and concatenation between the query features $\mathbf{Q}_{h}^{(i)}$ and the candidate features $\mathbf{F}_{s}^{(i)}$. The results in Table~\ref{tab:feature_fusion} show that the choice of fusion strategy affects both semantic understanding and the accuracy of candidate verification, providing empirical guidance for designing effective feature fusion methods.

\subsubsection{Ablation of Verification and Refinement Components}
Table~\ref{tab:reliable_feedback_ablation} evaluates the contributions of query supervision and refinement-control components in Bi-Refine. Both QS and RF individually improve the base model, indicating that supervision and reliability-aware feedback are independently beneficial. Adding AS on top of RF further improves all metrics, suggesting that adaptive stopping helps avoid unnecessary refinement during inference. Combining all three components achieves the best overall performance, reaching 70.45 R1@0.5, 54.00 R1@0.7, 68.82 mAP@0.5, 48.82 mAP@0.75, and 48.00 average mAP. These results demonstrate the complementary effects of query supervision and refinement control in Bi-Refine.

\subsubsection{Inference Efficiency}
Table~\ref{tab:inference_efficiency} reports latency for a fixed number of refinement loops. Moving from the baseline to Loop 1 adds 3.23 ms, while Loop 3 requires 30.64 ms in total, corresponding to a 15.4\% increase over the baseline. 

\subsubsection{Hyperparameter Sensitivity}
To investigate the sensitivity of the model to key hyperparameters, we conducted a series of experiments by adjusting several factors, including the number of loop iterations $N$, the number of queries $N_q$,  the loss balance weight $\lambda_{rel}$, and the context ratio $\gamma$. The results in Table~\ref{tab:horizontal_component_ablation} indicate that proper parameter settings balance semantic verification, localization accuracy, and efficiency, while extreme values can harm performance or increase computation. These ablations offer empirical guidance for hyperparameter selection.

\begin{figure*}[t]
\centering
  \includegraphics[width=0.8\textwidth]{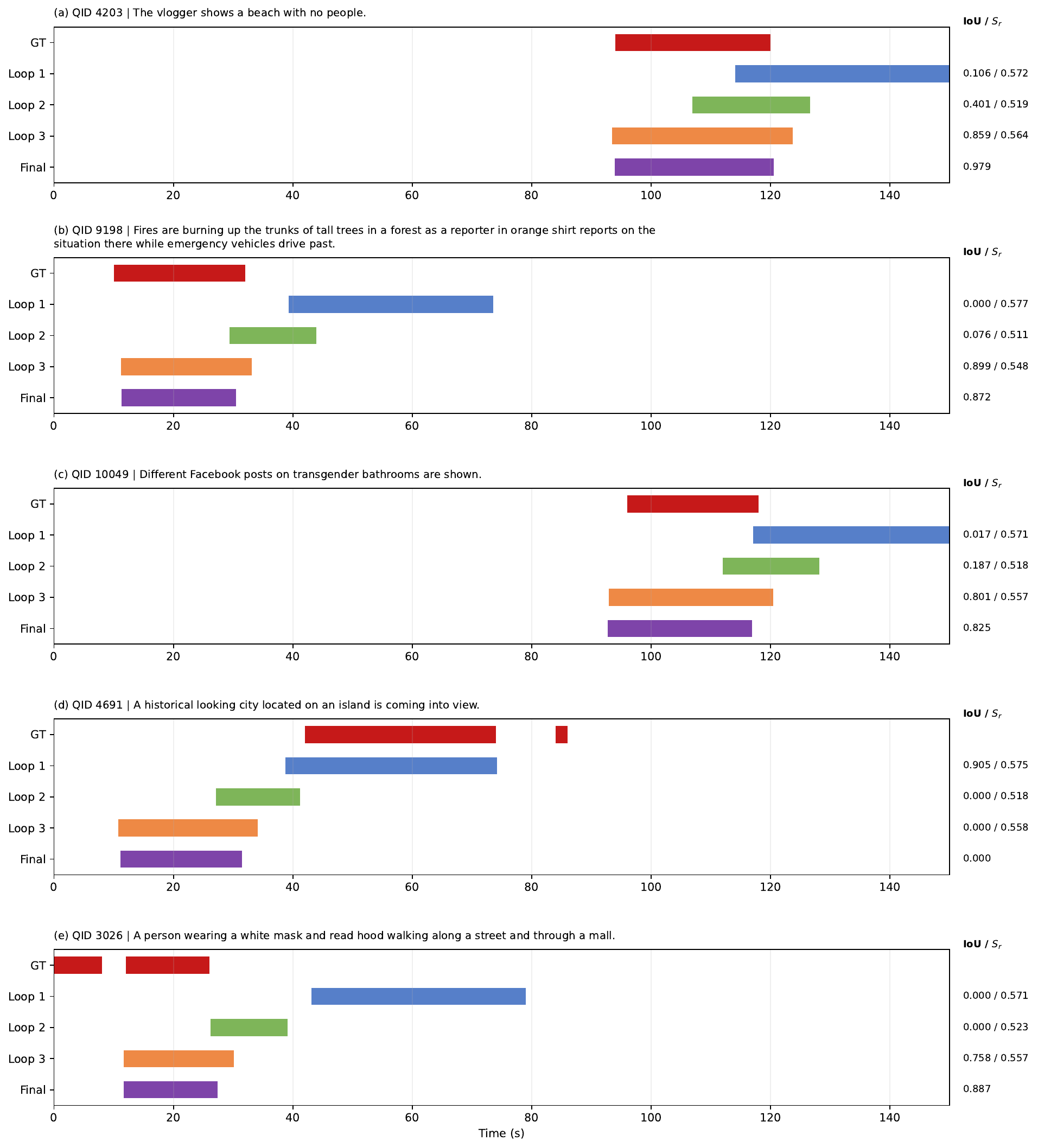}
\caption{Failure analysis of iterative feedback on the QVHighlights validation split. Case 4691 shows feedback-induced localization failure, whereas Case 3026 illustrates recovery from initially incorrect proposals. The remaining cases provide successful refinement references.}
  \label{fig:failure}
\end{figure*}

\begin{table*}[t]
\centering
\caption{Hyper-parameter ablations of the Bi-Refine on the QVHighlights validation split. Bold values indicate the best result in each group.}
\label{tab:horizontal_component_ablation}
\scriptsize
\setlength{\tabcolsep}{3.2pt}
\renewcommand{\arraystretch}{1.05}

\begin{minipage}[t]{0.49\textwidth}
\centering
\textit{(a) Number of loop iterations $N$}\par\vspace{2pt}
\begin{tabular}{c@{\hspace{5pt}}ccccc}
\toprule
$N$ & R@.5 & R@.7 & AP@.5 & AP@.75 & Avg. \\
\midrule
1 & 68.97 & 51.42 & 67.49 & 45.27 & 45.50 \\
2 & 68.32 & 52.58 & 66.36 & 45.82 & 44.91 \\
\rowcolor{gray!10}
3 & \textbf{70.45} & \textbf{54.00} & \textbf{68.82}
  & \textbf{48.82} & \textbf{48.00} \\
4 & 68.52 & 52.00 & 66.09 & 45.60 & 44.45 \\
\bottomrule
\end{tabular}
\end{minipage}
\hfill
\begin{minipage}[t]{0.49\textwidth}
\centering
\textit{(b) Number of queries $N_q$}\par\vspace{2pt}
\begin{tabular}{c@{\hspace{5pt}}ccccc}
\toprule
$N_q$ & R@.5 & R@.7 & AP@.5 & AP@.75 & Avg. \\
\midrule
5  & 68.06 & 50.84 & 64.59 & 43.29 & 43.08 \\
\rowcolor{gray!10}
10 & \textbf{70.45} & \textbf{54.00} & \textbf{68.82}
   & \textbf{48.82} & \textbf{48.00} \\
15 & 69.94 & 52.58 & 67.17 & 45.44 & 44.48 \\
20 & 68.90 & 51.94 & 66.60 & 45.70 & 45.01 \\
\bottomrule
\end{tabular}
\end{minipage}

\vspace{7pt}

\begin{minipage}[t]{0.49\textwidth}
\centering
\textit{(c) Relevance-loss weight $\lambda_{rel}$}\par\vspace{2pt}
\begin{tabular}{c@{\hspace{5pt}}ccccc}
\toprule
$\lambda_{rel}$ & R@.5 & R@.7 & AP@.5 & AP@.75 & Avg. \\
\midrule
0.5 & 68.32 & 50.52 & 65.18 & 44.87 & 43.41 \\
\rowcolor{gray!10}
1.0 & \textbf{70.45} & \textbf{54.00} & \textbf{68.82}
    & \textbf{48.82} & \textbf{48.00} \\
1.5 & 68.52 & 53.03 & 66.89 & 46.30 & 45.60 \\
2.0 & 68.39 & 50.84 & 66.79 & 45.73 & 44.66 \\
\bottomrule
\end{tabular}
\end{minipage}
\hfill
\begin{minipage}[t]{0.49\textwidth}
\centering
\textit{(d) Context ratio $\gamma$}\par\vspace{2pt}
\begin{tabular}{c@{\hspace{5pt}}ccccc}
\toprule
$\gamma$ & R@.5 & R@.7 & AP@.5 & AP@.75 & Avg. \\
\midrule
1.00 & 58.45 & 37.94 & 59.10 & 35.16 & 35.21 \\
1.25 & 67.16 & 50.77 & 65.58 & 44.27 & 43.85 \\
\rowcolor{gray!10}
1.50 & \textbf{70.45} & \textbf{54.00} & \textbf{68.82}
     & \textbf{48.82} & \textbf{48.00} \\
1.75 & 68.26 & 51.29 & 66.60 & 45.29 & 44.68 \\
\bottomrule
\end{tabular}
\end{minipage}
\end{table*}

\subsection{Qualitative Results} 
\noindent\textbf{Qualitative Analysis.}
Figure~\ref{fig:qualitative_loop} illustrates the evolution of temporal predictions throughout refinement. The initial proposals are often coarsely localized, while successive iterations may move them toward the ground-truth moments. For example, in cases (c) and (d), the IoU increases from 0.00 to 0.91 and from 0.11 to 0.98, respectively. Final denotes the ranked output from the last active state rather than an additional refinement iteration. Together with the failure cases below, these examples illustrate both the corrective behavior and non-monotonic nature of iterative semantic feedback.

\noindent\textbf{Failure case analysis.}
Figure~\ref{fig:failure} visualizes the feedback signals to analyze the failure behavior of iterative refinement. In Case 4691, Loop~1 closely matches the ground truth ($\mathrm{IoU}=0.905$), but subsequent feedback shifts the proposal to an irrelevant interval and eventually reduces the IoU to zero. The semantic relevance score nevertheless increases from Loop~2 to Loop~3, although both proposals have zero temporal overlap. This example shows that semantic relevance alone is not a reliable indicator of whether an update will improve localization and directly motivates the proposed feedback-reliability estimator. Case 3026 presents the complementary behavior: the first two loops miss the target, whereas later feedback redirects the proposal toward the correct moment. Thus, iterative verification can correct poor initialization, but applying its signals without reliability estimation can also cause over-correction. These diagnostic examples motivate reliability-aware feedback; they are not intended to claim that the revised model eliminates every failure mode.

\noindent\textbf{Qualitative results of Bi-Refine.} Figure~\ref{fig:qualitative} presents qualitative results of Bi-Refine on the QVHighlights dataset, compared with TR-DETR and LD-DETR. Bi-Refine produces temporal segments that align more closely with the ground truth. This improvement stems from the collaborative design and effective integration of the proposed modules, which enhance both temporal understanding and cross-modal alignment.

\FloatBarrier

\section{Discussion and Limitations}
\label{sec:discussion}

The framework distinguishes two quantities that are often conflated in iterative retrieval: semantic relevance describes the consistency of a candidate with the query, whereas feedback reliability describes whether using that evidence is expected to improve the next localization state. This distinction motivates regulating semantic feedback instead of treating every high relevance score as update confidence. The semantic-encoder comparison, controlled-feedback analysis, and failure cases are used to examine these design choices from complementary perspectives.

Bi-Refine still has two main limitations. First, because candidate-level semantic annotations are unavailable, the relevance branch uses Hungarian-matched labels as proxy supervision, and its scores may therefore partially reflect localization quality. Second, the improvement-based reliability objective remains sensitive to optimization and does not guarantee monotonic refinement for every sample. Future work may explore better-calibrated reliability targets and data-adaptive convergence criteria to improve stability across datasets and random initializations.

\section{Conclusion}

\label{sec:conclusion}


In this work, we presented Bi-Refine, a verification-guided refinement framework for moment retrieval. Rather than treating iterative retrieval as repeated localization alone, Bi-Refine explicitly separates candidate generation from semantic verification and establishes a closed interaction between the two processes. A key observation underlying the framework is that semantic relevance and feedback reliability are not equivalent: a candidate may appear semantically consistent with the query, while using its verification signal can still degrade temporal localization. By modeling these two quantities separately, Bi-Refine enables semantic evidence to guide refinement while reducing the risk of harmful updates. The consistent improvements on QVHighlights, TACoS, and Charades-STA, together with the component analyses and failure cases, demonstrate the effectiveness of this verification-and-refinement perspective across different moment retrieval settings.

Beyond the empirical gains, the framework provides a general way to reconsider iterative prediction in video-language grounding. Candidate-level verification can serve not only as an auxiliary semantic score, but also as an explicit intermediate reasoning signal for controlling subsequent localization. The distinction between semantic relevance and feedback reliability may therefore be useful for other iterative grounding or multimodal localization frameworks in which intermediate predictions are repeatedly reused for refinement. In this sense, Bi-Refine offers both a practical architecture and a diagnostic perspective for studying when semantic feedback should be trusted.

The current framework nevertheless has several limitations. First, candidate-level semantic annotations are unavailable in standard moment retrieval datasets, so the semantic relevance branch relies on Hungarian-matched candidates as proxy supervision. Consequently, the learned relevance score may still partially reflect localization quality rather than pure query-candidate semantic consistency. Second, the improvement-based reliability objective is sensitive to optimization and does not guarantee monotonic refinement for every sample, as also illustrated by the failure cases.

Future work will therefore investigate more direct and better-calibrated supervision for candidate-level semantic verification, more stable reliability objectives that can better predict the benefit of future refinement, and data-adaptive stopping criteria that are less sensitive to fixed thresholds. It would also be valuable to explore lightweight semantic encoders or knowledge distillation to reduce the computational overhead of LLM-based verification. More broadly, extending verification-guided refinement to other video-language grounding and multimodal localization tasks may help determine whether explicitly controlling the reliability of semantic feedback can provide a general mechanism for robust iterative inference.

\section*{Funding}
This work was supported in part by the National Natural Science Foundation of China under Grants 62572384, U24A20325, 62125305, and 62503384. Kun Xia was supported by the National Natural Science Foundation of China under Grant 62503384.

\section*{Declaration of competing interest}
The authors declare that they have no known competing financial interests or personal relationships that could have appeared to influence the work reported in this paper.



\bibliographystyle{elsarticle-num}
\bibliography{ref}

\end{document}